\PassOptionsToPackage{dvipsnames,table}{xcolor}
\documentclass[11pt, a4paper, logo, copyright, nonumbering]{map}

\usepackage[utf8]{inputenc} 
\usepackage[T1]{fontenc}    
\usepackage{hyperref}       
\usepackage{url}            
\usepackage{booktabs}       
\usepackage{amsfonts}       
\usepackage{nicefrac}       
\usepackage{microtype}      
\usepackage{fontawesome}
\usepackage{tcolorbox}
\tcbuselibrary{breakable}
\usepackage{multicol}
\usepackage[para,online,flushleft]{threeparttable}

\usepackage{natbib}

\usepackage{CJKutf8}

\usepackage{xargs}

\usepackage{multirow}

\usepackage{hyperref}
\usepackage{amsmath}
\usepackage{cleveref}
\usepackage{dsfont}

\usepackage{float} 
\usepackage{subcaption}
\usepackage{svg}
\usepackage{wrapfig}

\newcommand{\model}{\textsc{P2P}}
\newcommand{\method}{\textsc{P2P}}
\newcommand{\bench}{\textsc{P2Peval}}
\newcommand{\instruct}{\textsc{P2Pinstruct}}

\hypersetup{
    colorlinks=true,            
    linkcolor=blue,             
    filecolor=magenta,          
    urlcolor=cyan,              
    citecolor=purple,             
    pdftitle={Overleaf Example},
    pdfpagemode=FullScreen,
    }

\renewcommand{\today}{}
\newcommandx{\info}[2][1=]{\todo[linecolor=red,backgroundcolor=red!25,bordercolor=red,#1]{#2}}

\title{\centering P2P: Automated Paper-to-Poster Generation \\ and Fine-Grained Benchmark}

\author[1,2]{Tao Sun}
\author[1]{Enhao Pan}
\author[1]{Zhengkai Yang}
\author[1]{Kaixin Sui}
\author[2]{Jiajun Shi}
\author[2]{Xianfu Cheng}
\author[3]{Tongliang Li}
\author[1]{Wenhao Huang}
\author[1,2]{Ge Zhang}
\author[ ]{Jian Yang$^{\dagger}$}
\author[4]{Zhoujun Li$^{\dagger}$}

\affil[1]{ByteDance, China}
\affil[2]{M-A-P}
\affil[3]{College of Computer Science, Beijing Information Science and Technology University}
\affil[4]{Shenzhen Intelligent Strong Technology Co.,Ltd.}

\begin{abstract}
Academic posters are vital for scholarly communication, yet their manual creation is time-consuming. However, automated academic poster generation faces significant challenges in preserving intricate scientific details and achieving effective visual-textual integration.
Existing approaches often struggle with semantic richness and structural nuances, and lack standardized benchmarks for evaluating generated academic posters comprehensively. 
To address these limitations, we introduce \method{}, the first flexible, LLM-based multi-agent framework that generates high-quality, HTML-rendered academic posters directly from research papers, demonstrating strong potential for practical applications. \method{} employs three specialized agents—for visual element processing, content generation, and final poster assembly—each integrated with dedicated checker modules to enable iterative refinement and ensure output quality. To foster advancements and rigorous evaluation in this domain, we construct and release \instruct{}, the first large-scale instruction dataset comprising over 30,000 high-quality examples tailored for the academic paper-to-poster generation task. Furthermore, we establish \bench{}, a comprehensive benchmark featuring 121 paper-poster pairs and a dual evaluation methodology (Universal and Fine-Grained) that leverages LLM-as-a-Judge and detailed, human-annotated checklists. Our contributions aim to streamline research dissemination and provide the community with robust tools for developing and evaluating next-generation poster generation systems. The code is on the \url{https://github.com/multimodal-art-projection/P2P}.
\end{abstract}

\begin{document}
\begin{CJK*}{UTF8}{gbsn}

\maketitle

\newpage

\tableofcontents

\newpage

\section{Introduction}
\label{sec:intro}

Academic posters serve as a vital tool in scholarly communication, effectively distilling complex research into visually accessible formats for conferences and workshops, thereby fostering knowledge dissemination and collaborative engagement. However, manually creating these posters is often a time-consuming and skill-intensive process, particularly for early-career researchers who must balance content refinement with design proficiency. Automating academic poster generation presents a significant opportunity to streamline research dissemination, reducing barriers and enhancing accessibility for the broader academic community.

Existing approaches to poster generation primarily rely on template-based or rule-driven methods~\cite{xu2021neural}, which often struggle to capture the semantic richness and structural nuances of academic documents~\cite{qiang2019learning}, typically decomposing the task into isolated subtasks like content extraction~\cite{cheng2024sviptr}, panel attribute inference~\cite{huang2022layoutlmv3}, and layout generation~\cite{lin2024layoutprompter}. Although recent advances in multimodal large language models (MLLMs) and large language models (LLMs) show promise in understanding document structures and visual-textual relationships~\cite{jaisankar2024postdoc}, their application to academic poster generation remains limited due to insufficient quality control mechanisms and the absence of standardized benchmarks for systematic evaluation.

To address these challenges, we propose \method{}, the first flexible, LLM-based multi-agent framework designed for practical application in generating high-quality academic posters directly from research papers. \method{} employs three specialized agents—the Figure Agent for visual element processing, the Section Agent for content generation, and the Orchestrate Agent for final poster assembly, each paired with a dedicated checker module to enable iterative refinement and ensure output quality. This architecture streamlines the extraction and description of visual elements\cite{zhao2024doclayout}, the creation of structured content, and the seamless integration of these components into cohesive, HTML-rendered posters\cite{qiang2016learning}. 

Furthermore, to advance research in this domain, we release \instruct{}, the first large-scale instruction dataset comprising over 30,000 high-quality examples specifically tailored for the academic paper-to-poster generation task. Leveraging \instruct{}, we develop \texttt{Qwen3-P2P-8B}, which achieves powerful performance.

Concurrently, to facilitate rigorous assessment of the generated poster, we introduce \bench{}, a comprehensive benchmark comprising 121 paper-poster pairs with fine-grained annotations spanning diverse scientific disciplines. \bench{} implements a dual-pronged evaluation methodology using LLM-as-a-Judge: (1) a Universal Poster Evaluation assessing overall quality dimensions such as content fidelity, visual consistency, and layout effectiveness, and (2) a Fine-Grained Poster Evaluation that meticulously measures adherence to detailed, human-annotated checklists derived from official poster exemplars. We evaluate a total of 33 models. 

Our contributions are as follows: 

\begin{itemize}
    \item We propose \method{}, the first flexible, LLM-based multi-agent architecture for academic poster generation suitable for practical applications, which intelligently extracts and synthesizes content, and renders structured posters using HTML and CSS. 
    \item We introduce \instruct{}, the first large-scale instruction dataset specifically designed for academic poster generation. Comprising over 30K high-quality instruction-response pairs, \instruct{} captures the complete paper-to-poster transformation process.
    \item We establish \bench{}, a new benchmark for poster evaluation featuring a dual evaluation framework (Universal and Fine-Grained) that integrates human-annotated checklists and LLM-as-a-judge assessments for robust and multifaceted analysis.
\end{itemize}

\section{Methodology}
\label{sec:methodology}
This section details \method{} and \instruct{}, an automated multi-agent framework and training dataset designed for generating academic posters from research papers. 

\subsection{\method{}: Multi-agent for Paper-to-Poster Generation}

\begin{figure}[t]
    \centering
    \includegraphics[width=0.90\linewidth]{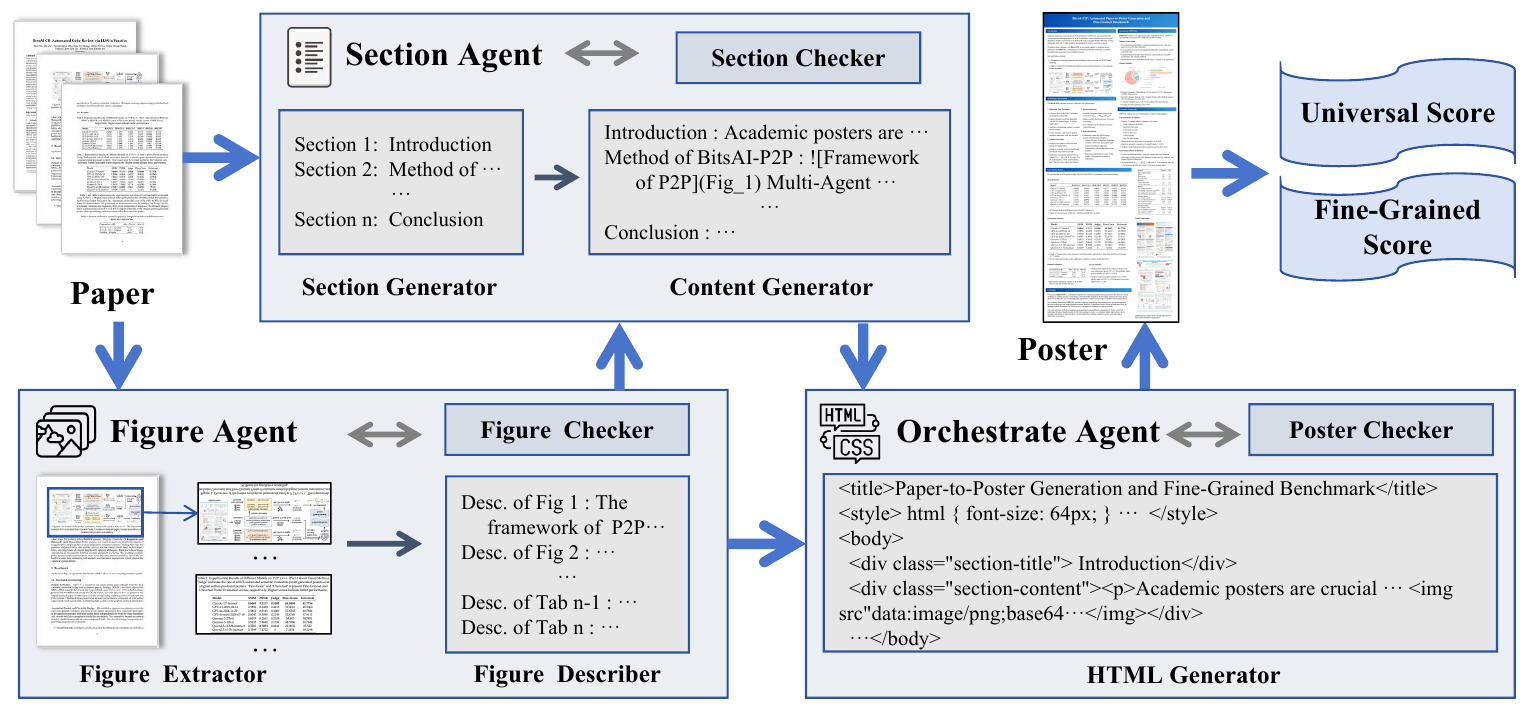}
    \caption{The multi-agent architecture of \method{}: papers are processed by the Figure Agent for extraction and description of visual elements, the Section Agent for structural and content generation, and the Orchestrate Agent for poster assembly and HTML rendering. Each agent employs checker modules and reflection loops for iterative enhancement.}
    \label{fig:system_architecture}
\end{figure}

As illustrated in Figure~\ref{fig:system_architecture}, the \method{} workflow is orchestrated by three collaborative agents: the Figure Agent, the Section Agent, and the Orchestrate Agent. Each agent operates in conjunction with a dedicated checker module to ensure its output quality, enabling an iterative refinement process.

\paragraph{Figure Agent.}
The Figure Agent is responsible for processing all visual elements within the input research paper. Its \textit{Figure Extractor} component employs DocLayout-YOLO~\citep{zhao2024doclayout}, a state-of-the-art document layout detection model, to extract figures and tables. Concurrently, the \textit{Figure Describer} identifies corresponding captions via spatial relation analysis. These components collaborate to synthesize semantic visual units by combining each extracted graphical component with its associated caption, yielding a set of described visual elements $\mathcal{F}_d = \{(v_1, c_1, \text{desc}_1), \dots, (v_n, c_n, \text{desc}_n)\}$. Here, $v_i$ denotes the visual, $c_i$ its original caption, and $\text{desc}_i$ a detailed description generated by an MLLM, $M_{figure}$. The \textit{Figure Checker} then validates this output by: (1) preventing duplicate extractions, (2) verifying the capture of all significant visual elements, and (3) confirming accurate visual-caption pairings. To ensure reliable pairings, an initial confidence threshold is applied to detected elements; this threshold is incrementally lowered if discrepancies arise between the counts of identified figures and captions, an iterative process repeated until sufficient alignment is achieved.

\paragraph{Section Agent.}
The Section Agent focuses on generating the textual content of the poster. Initially, the \textit{Section Generator} analyses the input paper ($D$) to dynamically infer a detailed structural schema ($S$) for the target poster. This schema delineates crucial sections (e.g., Introduction, Methods, Results) and their intended content focus. Subsequently, the \textit{Content Generator} synthesizes semantically coherent textual content for the poster, $P_{\text{poster\_text}}$, by utilizing the structural schema $S$, the original input paper $D$, and the detailed descriptions and indices of visual elements $\mathcal{F}_d$ provided by the Figure Agent. This textual content generation can be formally described as:
$ P_{\text{poster\_text}} = \mathcal{M}_{\text{text}}(D, S, \mathcal{F}_d) $,
where $\mathcal{M}_{\text{text}}$ is a LLM specialized in text generation. $\mathcal{M}_{\text{text}}$ employs prompts not only to generate text but also to strategically integrate Markdown-style references to figure indices from $\mathcal{F}_d$ at optimal textual positions, ensuring contextual relevance and visual-textual alignment. The \textit{Section Checker} scrutinizes the generated $P_{\text{poster\_text}}$ for: (1) coherence and logical flow, (2) completeness in covering core contributions, (3) faithfulness to the original paper's findings, and (4) correct and relevant referencing of visual elements. If inadequacies are detected, a reflection loop initiates a revision of the section structure or content by the Section Agent.

\paragraph{Orchestrate Agent.}
The Orchestrate Agent integrates the visual and textual components into a cohesive and professionally formatted poster. The \textit{HTML Generator} utilizes the Markdown-formatted text $P_{\text{poster\_text}}$ from the Section Agent and the actual visual elements (images/tables $\mathcal{F}_v$, where each figure is additionally provided with its width, height, and aspect ratio as supplementary information) extracted by the Figure Agent, to produce the poster in HTML and CSS. 
The Orchestrate Agent deliberately omits original captions from $\mathcal{F}_d$ in the final embedded visuals to improve visual clarity and maintain a concise academic presentation. 
The rendering process adheres to three principles:
(1) Content-Structure Decoupling: Decouple semantics from presentation via modular CSS. 
(2) Institutional Identity Alignment: Customize color schemes to align with the logo of institution or conference. 
(3) Responsive and Balanced Layout Generation: Use CSS flexbox for adaptive column structures and whitespace optimization. 
The \textit{Poster Checker} evaluates the rendered poster for layout aesthetics and structural integrity, triggering iterative adjustments (via reflection) to resolve issues like unbalanced spacing or misaligned elements until the design meets professional standards.

\begin{figure}[t]
    \centering
    \includegraphics[width=0.67\linewidth]{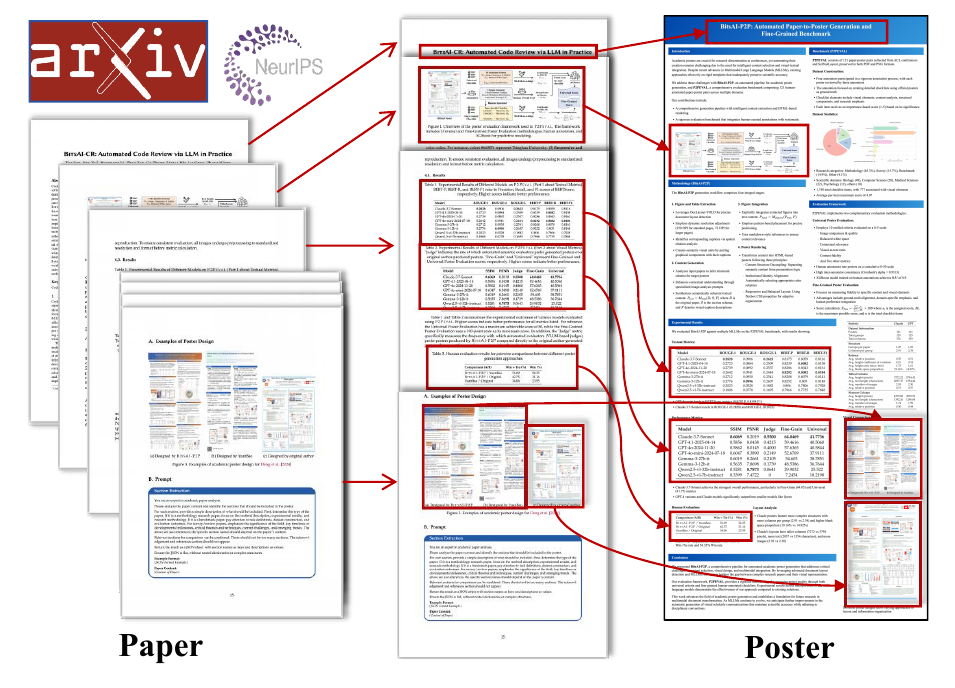}
    \caption{An example of the paper-to-poster transformation achieved by \method{}, showing direct correspondences between elements in the input paper (left) and the generated academic poster (right).}
    \label{fig:paper_to_poster_example}
\end{figure}

Figure~\ref{fig:paper_to_poster_example} illustrates the core transformation process facilitated by \method{}. On the left, a multi-page academic research paper, sourced from repositories such as arXiv or conference proceedings like NeurIPS, serves as the input. On the right, the corresponding academic poster, generated by \method{}, is displayed. The red arrows explicitly map key elements from the original paper, such as the title, specific figures, and sections, to their respective locations and representations in the final poster.

\subsection{\instruct{}: A Large-Scale Instruction Dataset for Paper-to-Poster Generation}

The \instruct{} dataset is derived from the \method{} to support training of models for poster generation. Following \method{}, we collect 30,460 high-quality instruction-response pairs spanning the complete poster generation workflow. For visual element processing, we prompt Claude to generate 16,848 figure-description pairs through the Figure Describer component, yielding descriptive texts averaging 192 tokens per visual element. For textual content generation, we collect 13,612 instruction-response pairs from the Section Generator, Content Generator, and HTML Generator components. These examples average over 3,300 tokens per response, demonstrating the complexity and richness of the generated content.

\section{\bench{}: A Fine-grained Benchmark for Poster Evaluation}
\label{sec:benchmark}

\begin{figure}[t]
    \centering
    \includegraphics[width=0.98\linewidth]{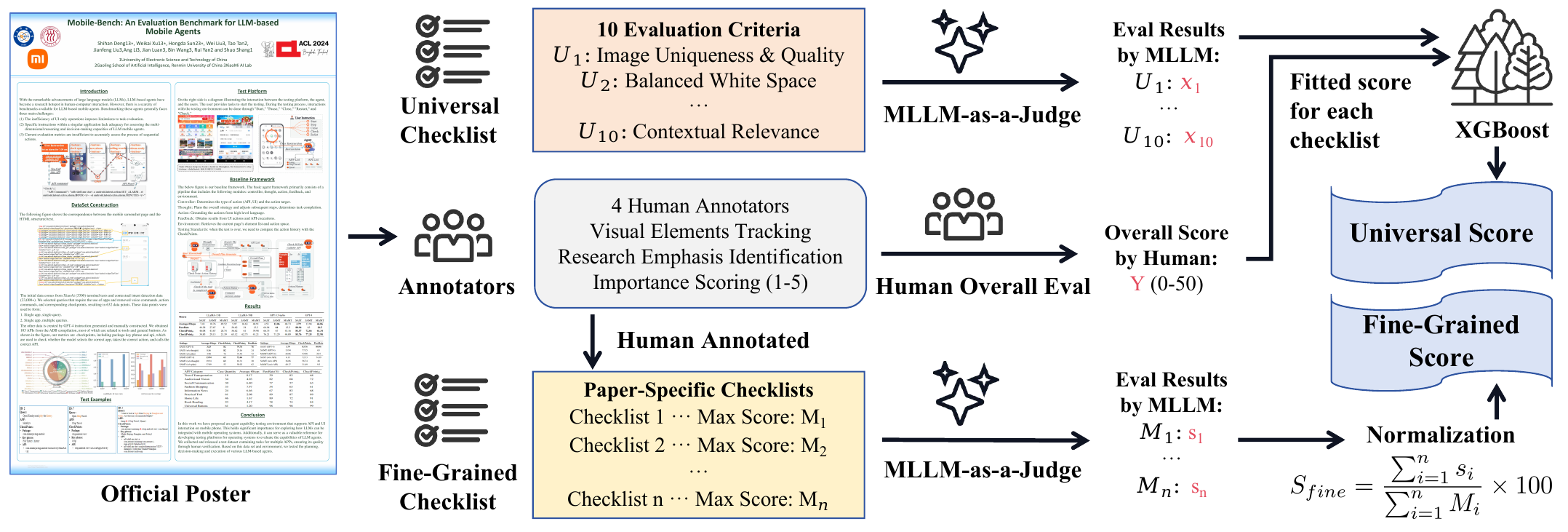}
    \caption{Overview of the poster evaluation framework used in \bench{}. \bench{} includes Universal and Fine-Grained Poster Evaluation, human annotators, and XGBoost for scoring.}
    \label{fig:eval}
\end{figure}

As shown in Fig~\ref{fig:eval}, we present a benchmark called \bench{} for evaluating academic posters.

\subsection{Benchmark Constructing}
\label{sections/dataset_collection}

\paragraph{Annotation Process and Checklist Design.} 
\label{par:checklist}
We establish a rigorous annotation protocol to create fine-grained evaluation criteria for poster quality assessment. Four annotators participate in the annotation process, with each poster being independently reviewed by three annotators and a fourth individual assigned to verify the annotations. 
The annotation focuses on creating detailed checklists using official posters as ground truth. Our checklist design incorporates the following elements: 
\textbf{(1) Visual Elements}: Each figure or table present in the official poster constitutes an individual checklist item, evaluated based upon its presence and accurate representation. 
\textbf{(2) Content Analysis}: Each visual element is assessed regarding its textual consistency with the original poster and its visual prominence within poster layout. 
\textbf{(3) Structural Components}: Annotators identify critical sections such as task definitions, experimental methodologies, and research conclusions within each poster panel. 
\textbf{(4) Research Emphasis}: Essential research findings, methodological details, and explicitly highlighted motivations (often noted by bold or prominent placement) form individual checklist items. 
\textbf{(5) Scoring System}: Each checklist item receives an importance-based score ranging from 1 to 5—minor details rated as 1, core elements as 3, and critical components central to the paper as 5. 
All checklist annotations, including unique paper identification, detailed evaluation criteria, reference figures (when applicable), and established maximum scores, are documented in YAML format.

\paragraph{Dataset Collection and Statistics.}

\begin{figure}[t]
    \centering
    \includegraphics[width=0.85\linewidth]{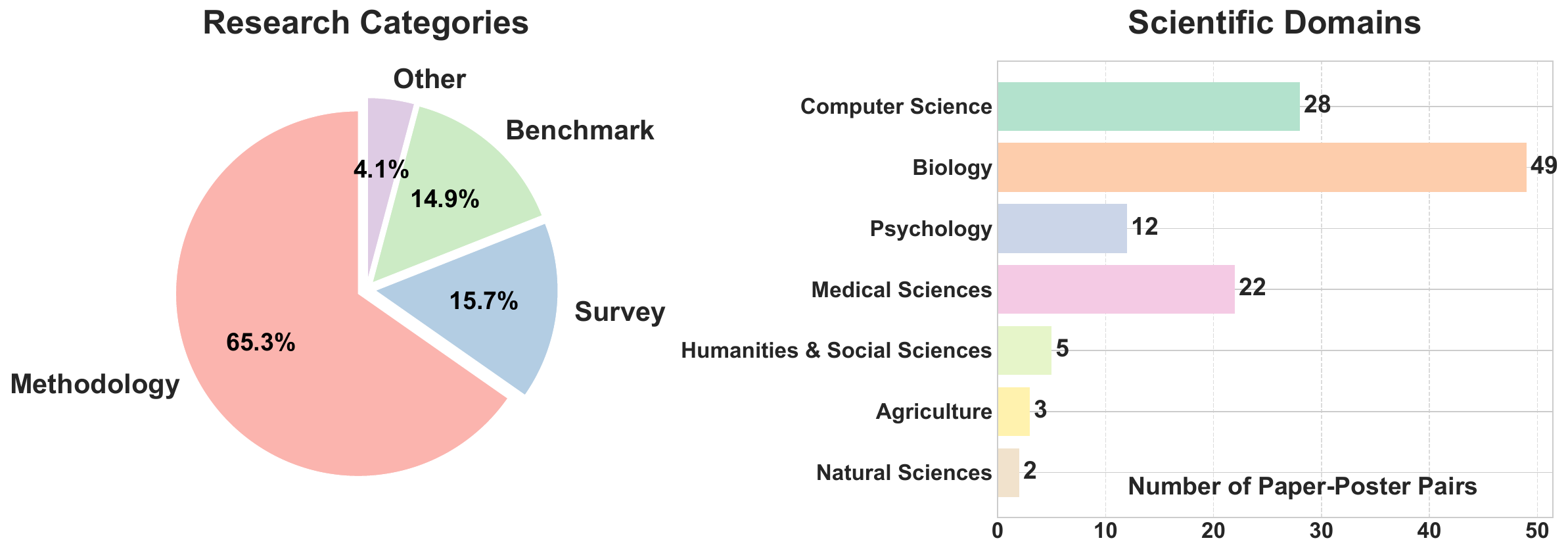}
    \caption{Distribution of \bench{}.}
    \label{fig:dataset}
\end{figure}

\bench{} consists of 121 paper-poster pairs collected from the ACL conference series (from 2022 to 2024) under CC4.0 license and from SciPostLayout~\citep{wang2024scipostlayout}, which contains posters from F1000Research under the CC-BY license. For each pair, \bench{} preserves the original research paper in PDF format and the corresponding academic poster in both PDF and PNG formats. This dual-format preservation enables comprehensive evaluation of both textual content and visual layout while maintaining high-quality vector graphics and text information. 
As shown in the Fig~\ref{fig:dataset}, \bench{} encompasses a broad range of research categories and disciplines. The annotation process results in 1738 checklist items, with 775 associated with visual elements. The scoring system yields an average per-item maximum score of 4.07. 

\subsection{Poster Evaluation Framework}
\label{subsec:evaluate}

Our evaluation pipeline consists of two complementary methodologies: Universal Poster Evaluation and Fine-Grained Poster Evaluation. We describe these two evaluation methods in detail below.

\subsubsection{Universal Poster Evaluation}
\label{subsubsec:universal_eval}

Universal Poster Evaluation employs a unified set of evaluation criteria, each evaluated independently on a discrete scale ranging from 0 to 5. These universal criteria ($U_1$ through $U_{10}$) include:

\begin{multicols}{2}
\begin{itemize}
    \item $U_1$: Authorship and Title Accuracy
    \item $U_2$: Image Uniqueness and Quality
    \item $U_3$: Balanced White Space
    \item $U_4$: Contextual Relevance
    \item $U_5$: Optimal Visual-to-Text Ratio
    \item $U_6$: Dimension Appropriateness
    \item $U_7$: Visual Consistency
    \item $U_8$: Content Fidelity
    \item $U_9$: Information Flow Logic
    \item $U_{10}$: Self-Contained Explanation
\end{itemize}
\end{multicols}

To rigorously validate the LLM-based evaluation effectiveness, three human annotators independently rate posters on a cumulative 0–50 scale based on the above universal criteria. This human evaluation covers all original posters and our \method{} outputs, excluding multi-agent approaches (without checker and reflection mechanisms). 
We utilize both powerful models like GPT-4o and lighter models such as Qwen-VL-2.5-32B, ensuring the trained annotators are exposed to diverse samples to enhance generalizability. In total, we accumulate 1,701 independent annotation scores.

To train the scoring model, we utilize XGBoost, a robust gradient boosting framework, to model the nonlinear interactions between universal evaluation criteria scores and corresponding human annotations. Specifically, the XGBoost model undergoes training with 10-fold cross-validation and utilizes 200 trees. The resulting predictive model exhibits strong performance, achieving an $R^{2}$ of 0.92, thus validating the reliability and effectiveness of our Universal Poster Evaluation pipeline. 
Additionally, we experiment with other methods, including Ordinary Least Squares($R^2=0.66$), Random Forest($R^2=0.83$), and various regularization techniques($R^2=0.89$); however, these approaches yield suboptimal performance compared to XGBoost.

\subsubsection{Fine-Grained Poster Evaluation}
\label{subsubsec:fine_grained_eval}

Complementing the universal evaluation, we design a Fine-Grained Poster Evaluation pipeline, focused explicitly on measuring each generated poster's fidelity to specific content and visual elements in official academic posters. The fine-grained evaluations directly employ detailed checklists produced in our annotation workflow (Section~\ref{par:checklist}). Each checklist item's maximum score is consensus-derived from multiple annotators, ranging from minor visual elements assigned a score of 1 to core research components scored at 5, reflecting their relative importance. 
This human-centered approach ensures that the scoring system inherently embodies human preferences and domain expertise. 

We formally define the final fine-grained evaluation score $S_{fine}$ as $S_{fine} = \frac{\sum_{i=1}^{n} s_i}{\sum_{i=1}^{n} M_i} \times 100$, 
where \( S_{fine} \) is the normalized fine-grained evaluation score on a 0–100 scale, \( s_i \) is the assigned score for the \( i^{th} \) checklist item represented in the generated poster, \( M_i \) denotes the corresponding maximum possible score for that item, and \( n \) signifies the total number of checklist items.  Consequently, the Fine-Grained Poster Evaluation score comprehensively assesses a generated poster's capability to faithfully preserve the original research's essential content and visual priorities. By emphasizing explicit fidelity to the original author's intended communication goals rather than generic quality alone, the approach enables clear comparative analyses across diverse poster generation methodologies.

\section{Experiments and Analysis}

\begin{table}[t!]
\centering
\caption{Experimental results of different models on \bench{}. Higher scores indicate better.}
\resizebox{1\textwidth}{!}{
\begin{threeparttable}[b]
\begin{tabular}{@{}lcccccccc@{}}
\toprule
\textbf{Model} & \textbf{Size} & \textbf{ROUGE-1} & \textbf{ROUGE-2} & \textbf{ROUGE-L} & \textbf{BERT\tnote{1}} & \textbf{Judge\tnote{2}} & \textbf{FineGrain\tnote{3}} & \textbf{Universal\tnote{4}} \\ \midrule
\multicolumn{9}{c}{\textbf{Closed-Source Models}} \\ \midrule
\multicolumn{1}{l|}{Claude-3.7-Sonnet} & \faLock & 0.2745 & 0.0830 & 0.2527 & \multicolumn{1}{c|}{0.8109} & 0.5537 & 65.3962 & \textbf{37.2474} \\
\multicolumn{1}{l|}{Claude-3.7-Sonnet\tnote{R}} & \faLock & 0.2734 & 0.0848 & 0.2516 & \multicolumn{1}{c|}{0.8111} & \textbf{0.6281} & \textbf{65.8848} & 35.5062 \\
\multicolumn{1}{l|}{Claude-3.5-Sonnet} & \faLock & 0.2367 & 0.0615 & 0.2185 & \multicolumn{1}{c|}{0.8081} & 0.2810 & 47.7385 & 30.2544 \\
\multicolumn{1}{l|}{GPT-4.1-2025-04-14} & \faLock & 0.2459 & 0.0685 & 0.2281 & \multicolumn{1}{c|}{0.8113} & 0.4793 & 60.2879 & 34.4700 \\
\multicolumn{1}{l|}{GPT-4.1-mini-2025-04-14} & \faLock & 0.2616 & 0.0741 & 0.2407 & \multicolumn{1}{c|}{0.8125} & 0.3388 & 55.3493 & 31.0697 \\
\multicolumn{1}{l|}{GPT-4.1-nano-2025-04-14} & \faLock & 0.2169 & 0.0557 & 0.1990 & \multicolumn{1}{c|}{0.8070} & 0.2066 & 41.3446 & 27.7149 \\
\multicolumn{1}{l|}{GPT-4o-2024-11-20} & \faLock & 0.2395 & 0.0668 & 0.2217 & \multicolumn{1}{c|}{0.8114} & 0.4959 & 55.4380 & 34.3888 \\
\multicolumn{1}{l|}{GPT-4o-mini-2024-07-18} & \faLock & 0.2362 & 0.0732 & 0.2198 & \multicolumn{1}{c|}{\textbf{0.8167}} & 0.2314 & 48.8879 & 30.8409 \\
\multicolumn{1}{l|}{OpenAI-o1\tnote{R}} & \faLock & 0.2385 & 0.0611 & 0.2200 & \multicolumn{1}{c|}{0.8088} & 0.3103 & 56.8504 & 34.1659 \\
\multicolumn{1}{l|}{Seed1.5-VL\tnote{R}} & \faLock & 0.2160 & 0.0539 & 0.2026 & \multicolumn{1}{c|}{0.8041} & 0.4050 & 62.4702 & 33.9840 \\
\multicolumn{1}{l|}{Seed-Thinking-v1.5\tnote{RT}} & \faLock & 0.2357 & 0.0701 & 0.2210 & \multicolumn{1}{c|}{0.8113} & 0.4711 & 61.9632 & 34.6882 \\
\multicolumn{1}{l|}{Seed-Thinking-v1.5-m\tnote{R}} & \faLock & 0.2493 & 0.0767 & 0.2315 & \multicolumn{1}{c|}{0.8116} & 0.3719 & 57.1457 & 33.2461 \\
\multicolumn{1}{l|}{Doubao-1.5-vision-pro} & \faLock & 0.2586 & 0.0849 & 0.2409 & \multicolumn{1}{c|}{0.8089} & 0.0354 & 45.9282 & 14.0841 \\
\multicolumn{1}{l|}{YuanBao\tnote{5}} & \faLock & - & - & - & \multicolumn{1}{c|}{-} & 0.0083  & 57.8677 & 31.5754  \\ \midrule
\multicolumn{9}{c}{\textbf{6B+ Models}} \\ \midrule
\multicolumn{1}{l|}{InternVL3} & 8B & 0.1980 & 0.0618 & 0.1847 & \multicolumn{1}{c|}{0.7994} & 0.0776 & 33.3900 & 22.2245 \\
\multicolumn{1}{l|}{Qwen3\tnote{T}} & 8B & 0.2563 & 0.0859 & 0.2373 & \multicolumn{1}{c|}{0.8152} & 0.1835 & 45.0272 & 28.8107 \\
\multicolumn{1}{l|}{Qwen3\tnote{RT}} & 8B & 0.2231 & 0.0619 & 0.2082 & \multicolumn{1}{c|}{0.8125} & 0.2545 & 53.6611 & 32.4912 \\
\multicolumn{1}{l|}{Qwen2.5-VL} & 7B & 0.1090 & 0.0414 & 0.1020 & \multicolumn{1}{c|}{0.7645} & 0.0083 & 13.7417 & 13.0597 \\
\midrule \multicolumn{9}{c}{\textbf{12B+ Models}} \\ \midrule
\multicolumn{1}{l|}{Gemma-3} & 12B & 0.2411 & 0.0764 & 0.2250 & \multicolumn{1}{c|}{0.8096} & 0.0940 & 46.7903 & 27.3686 \\
\multicolumn{1}{l|}{InternVL3} & 14B & 0.2437 & 0.0736 & 0.2253 & \multicolumn{1}{c|}{0.8132} & 0.0756 & 45.5513 & 25.6062 \\ \midrule
\multicolumn{9}{c}{\textbf{27B+ Models}} \\ \midrule
\multicolumn{1}{l|}{Gemma-3} & 27B & 0.2500 & 0.0794 & 0.2346 & \multicolumn{1}{c|}{0.8133} & 0.2857 & 50.8931 & 28.7410 \\
\multicolumn{1}{l|}{Gemma-3\tnote{T}} & 27B & 0.2536 & 0.0853 & 0.2372 & \multicolumn{1}{c|}{0.8132} & 0.2417 & 52.1716 & 28.5901 \\
\multicolumn{1}{l|}{InternVL3} & 38B & 0.2440 & 0.0756 & 0.2258 & \multicolumn{1}{c|}{0.8143} & 0.2333 & 52.6634 & 29.5850 \\
\multicolumn{1}{l|}{Qwen3\tnote{RT}} & 3/30B & 0.2270 & 0.0637 & 0.2125 & \multicolumn{1}{c|}{0.8120} & 0.2562 & 52.2125 & 31.1930 \\
\multicolumn{1}{l|}{Qwen3\tnote{RT}} & 32B & 0.2314 & 0.0659 & 0.2168 & \multicolumn{1}{c|}{0.8090} & 0.1736 & 46.0383 & 28.9479 \\
\multicolumn{1}{l|}{Qwen2.5-Coder\tnote{T}} & 32B & 0.2666 & 0.0949 & 0.2487 & \multicolumn{1}{c|}{\textbf{0.8167}} & 0.3884 & 55.9441 & 32.7935 \\
\midrule
\multicolumn{9}{c}{\textbf{72B+ Models}} \\ \midrule
\multicolumn{1}{l|}{Deepseek-R1\tnote{RT}} & 37/671B & 0.1927 & 0.0461 & 0.1795 & \multicolumn{1}{c|}{0.8015} & 0.5333 & 62.5013 & 33.9701 \\
\multicolumn{1}{l|}{Deepseek-V3\tnote{T}} & 37/671B & 0.2371 & 0.0739 & 0.2232 & \multicolumn{1}{c|}{0.8124} & 0.5041 & 59.6805 & 33.6045 \\ 
\multicolumn{1}{l|}{InternVL3} & 78B & 0.2424 & 0.0789 & 0.2245 & \multicolumn{1}{c|}{0.8152} & 0.2773 & 51.2962 & 28.9230 \\
\multicolumn{1}{l|}{Qwen3\tnote{RT}} & 22/235B & 0.2278 & 0.0625 & 0.2141 & \multicolumn{1}{c|}{0.8077} & 0.3967 & 53.7927 & 31.4551 \\
\multicolumn{1}{l|}{Qwen2.5-VL} & 72B & 0.2577 & 0.0909 & 0.2400 & \multicolumn{1}{c|}{0.8148} & 0.2833 & 55.7929 & 32.3105 \\

\rowcolor{green!8} \multicolumn{1}{l|}{Qwen3-P2P\tnote{T6}} & 8B & \textbf{0.2882} & \textbf{0.0955} & \textbf{0.2675} & \multicolumn{1}{c|}{0.8135} & 0.4587 & 57.6622 & 32.4996 \\
\rowcolor{green!8} \multicolumn{1}{l|}{Qwen2.5-VL-P2P\tnote{7}} & 7B & 0.1939 & 0.0609 & 0.1797 & \multicolumn{1}{c|}{0.7926} & 0.3140 & 37.3078 & 25.0337 \\

\bottomrule
\end{tabular}
\begin{tablenotes}
\item [R] Reasoning/Thinking Mode.
\item [T] Because they are text-only LLMs, we use \texttt{Claude-3.7-Sonnet} as the provider of Figure Describer.
\item [1] F1 scores of BERTScore.
\item [2] The rate at which LLM-as-a-Judger prefer generated posters over original author-produced posters.
\item [3] Scores of Fine-Grained Poster Evaluation.
\item [4] Scores of Universal Poster Evaluation.
\item [5] Posters generated by Tencent's AI application called YuanBao.
\item [6] Our model, built on \texttt{Qwen3-8B}, is fine-tuned using \instruct{}. 
\item [7] Our model, built on \texttt{Qwen2.5-VL-8B}, is fine-tuned using \instruct{}. 
\end{tablenotes}
\end{threeparttable}}
\label{tab:main_results_part1}

\end{table}

\subsection{Experimental Setup}
\label{subsec:setup}

We conduct comprehensive experiments to evaluate \model{} against several MLLMs on \bench{}. Specifically, we compare these different model series: 
GPT\cite{achiam2023gpt}, Claude\cite{Claude_3}, Doubao\cite{seed2025seed}, Qwen\cite{bai2025qwen2}, InternVL\cite{chen2024internvl}, Gemma\cite{team2025gemma}, Deepseek\cite{guo2025deepseek,liu2024deepseek}. All models are configured with the temperature of $1$ and the maximum output token length of 8000 to ensure fair comparison while maintaining generation diversity. 
Additionally, we include in our evaluation poster images generated by Tencent's AI application, YuanBao (\url{https://yuanbao.tencent.com/}), which directly produces academic posters in image format and Chinese. 
We also fine-tune our model \texttt{Qwen3-P2P} and \texttt{Qwen2-VL-P2P} using \instruct{} with a learning rate of 5×10$^{-5}$ for 3 epochs, employing AdamW\cite{loshchilov2017decoupled} and a maximum sequence length of 8000.

\subsection{Evaluation Metrics}

Beyond the human-validated Universal Poster Evaluation (max 50) and Fine-Grained Poster Evaluation (max 100) using \texttt{GPT-4o} in \bench{}, we supplement analysis with objective metrics: (1) ROUGE \citep{lin2004rouge}, which measure n-gram overlap between generated and reference poster content, thus capturing lexical similarity. (2) BERTScore \citep{zhang2019bertscore}, which leverages contextual embeddings to assess semantic similarity. During evaluation, all image links are removed from the text to ensure fair comparison of purely textual content. And the ``Judge'' metric reports how frequently VLLM-based automated evaluators prefer \method{}'s posters over original author-created versions.

\subsection{Results and Analysis}

\paragraph{Main Results.}
Table~\ref{tab:main_results_part1} summarizes the experimental outcomes of various models evaluated using \bench{}. Our analysis reveals several key findings:
\textbf{(1) Closed- vs. Open-source Models:} Closed-source models, notably \texttt{Claude-3.7-Sonnet}, achieve superior performance in qualitative assessments like the Universal and Fine-Grained evaluation. And leading open-source models such as \texttt{Deepseek-R1}(using \texttt{Claude} as the provider of Figure Describer), demonstrate strong competitiveness. 
\textbf{(2) Impact of Reasoning Capabilities:} Models employing reasoning or thinking modes such as \texttt{Claude-3.7-Sonnet} and \texttt{Qwen3} consistently show enhanced performance, especially in the Fine-Grained evaluation. This suggests that advanced reasoning aids in generating outputs that are more aligned with human preferences and detailed content requirements. 
\textbf{(3) Efficacy of \instruct{}:} Fine-tuning models on \instruct{} dataset yields substantial improvements. The \texttt{Qwen3-P2P-8B} achieves the highest ROUGE scores across all evaluated models, significantly outperforming its base version and even leading closed-source models in these lexical metrics. It also demonstrates considerable gains in FineGrain and Universal scores over \texttt{Qwen3}. Likewise, \texttt{Qwen2.5-VL-P2P-7B} shows enhancements across all metrics compared to its base model. These results underscore the value of \instruct{}. 
\textbf{(4) Supplemental Observations:} A divergence among evaluation criteria is also evident—excellence in lexical overlap (ROUGE) does not uniformly correlate with detailed fidelity (FineGrain), emphasizing the comprehensive nature of \bench{}. The strong performance of text-only models utilizing \texttt{Claude} for figure description points to the effectiveness of modular, hybrid approaches in this complex generation task.

\begin{table}[t!]
 \begin{minipage}[t]{0.6\textwidth}
  \centering
    \makeatletter\def\@captype{table}\makeatother\caption{Results of pairwise human preference evaluations.}
     \label{tab:human_eval}
\begin{tabular}{ccc}
    \toprule
\textbf{\begin{tabular}[c]{@{}c@{}}Comparison\\ (A vs. B)\end{tabular}} & \textbf{\begin{tabular}[c]{@{}c@{}}Preferred or Tied \\ (\%)\end{tabular}} & \textbf{\begin{tabular}[c]{@{}c@{}}Preferred \\ (\%)\end{tabular}} \\
\midrule
    \model{} / YuanBao & 83.05 & 54.35 \\
    \model{} / Original & 57.63 & 35.59 \\
    YuanBao / Original & 20.34 & 12.40\\
    \bottomrule
    \end{tabular}
  \end{minipage}
  \begin{minipage}[t]{0.4\textwidth}
   \centering
\makeatletter\def\@captype{table}\makeatother\caption{Performance comparison of \model{} across different output format.}
\label{tab:Output}
\begin{tabular}{@{}c|cc@{}}
\toprule
\textbf{Output} & \textbf{FineGrain} & \textbf{Universal} \\ \midrule
HTML & \textbf{65.3962} & \textbf{37.2474} \\
SVG & 52.7408 & 30.6648 \\
LaTex & 56.8756 & 25.2585 \\ \bottomrule
\end{tabular}
\end{minipage}
\end{table}

\paragraph{Analysis of Human Preference Evaluation.}
To complement our \bench{}, we conduct pairwise human preference evaluations, the results of which are presented in Table\ref{tab:human_eval}. 
Participants compare posters generated by \model{} using \texttt{Claude-3.7-Sonnet}, Tencent's YuanBao, and the original author-created posters. The "Preferred or Tied (\%)" and "Strictly Preferred (\%)" quantify the proportion of instances where method A is judged superior or equivalent to, and strictly superior to, method B, respectively. 
The results demonstrate a clear preference for \model{}-generated posters over those from YuanBao. Notably, \model{} also shows competitive performance against original posters, suggesting its capability to produce posters of superior quality in a significant number of cases.

\paragraph{Analysis of Output Format Selection.} 
Our investigation of different output formats using \texttt{Claude-3.7-Sonnet} reveals HTML as the optimal medium for academic posters. As documented in Table~\ref{tab:Output}, HTML-based poster outputs consistently outperform SVG and LaTeX alternatives across both fine-grained and universal metrics. 
The inherent flexibility of HTML and CSS for layout structuring and content decoupling, coupled with the robust rendering capabilities of modern browsers, contributes to this performance. 
Furthermore, our experiments suggest that current LLMs exhibit greater proficiency in HTML code generation compared to equivalent SVG or LaTeX implementations, resulting in fewer rendering errors or structural inconsistencies in the final poster artifacts. 

\begin{table}[t!]
\centering
\caption{Ablation study results by \texttt{Claude-3.7-Sonnet}.}
\label{tab:ablation}
\resizebox{0.6\textwidth}{!}{
\begin{tabular}{@{}ccc|cc@{}}
\toprule
\textbf{Mutli Agent} & \textbf{Figure Describer} & \textbf{Reflection} & \textbf{FineGrain} & \textbf{Universal} \\ \midrule
\checkmark & \checkmark & \checkmark & \textbf{65.3962} & \textbf{37.2474} \\
\checkmark & \checkmark &  & 64.4556 & 34.2229 \\
\checkmark &  & \checkmark & 63.7388 & 35.1107 \\
\checkmark &  &  & 63.5806 & 33.1458 \\
 &  &  & 60.7233 & 34.2554 \\ \bottomrule
\end{tabular}}
\end{table}

\paragraph{Ablation Study.} The results of ablation study of \method{} in Table~\ref{tab:ablation} demonstrate that the full system consistently outperforms reduced configurations. 
When reflection mechanisms (implemented through checker modules) are removed, we observe a moderate decline in universal metrics, suggesting these iterative feedback loops enhance overall poster quality and aesthetic coherence. 
Similarly, ablating the Figure Describer component, which transforms visual elements into textual descriptions, results in performance degradation. This indicates that directly feeding raw images to MLLMs for content integration can be less effective than providing them with semantically rich textual summaries. These descriptions appear to reduce the interpretative burden on the MLLMs and facilitate a more accurate contextualization of visual information within the poster. 
Removing all specialized components (resulting in a direct paper-to-poster pipeline without intermediate processing) leads to the greatest performance drop in fine-grained evaluation. 
This confirms our hypothesis that poster generation benefits significantly from modularized, specialized processing that mimics the distinct cognitive steps humans undertake when creating posters from research papers.

\paragraph{Analysis of Layout without Reflection.} A comparative analysis of poster layouts generated by Claude and GPT models, summarized in Table~\ref{tab:layout_comparison}, reveals distinct structural tendencies inherent in content segmentation and spatial organization to each model when operating without reflection mechanisms. 
Claude-generated posters typically exhibit a more fragmented structure, utilizing a greater number of columns. These layouts also demonstrate a tendency towards imbalanced spatial distribution, with taller content often concentrated towards the right and greater variability in column heights. This often results in a higher proportion of blank space, suggesting less efficient spatial utilization. 
In contrast, GPT-generated posters generally present more uniform and compact layouts. These findings suggest challenges in achieving consistent content allocation across the poster layout, a critical aspect for visual appeal and readability in academic posters.

\begin{table}[t!]
\centering
\caption{Comparison of layout statistics in posters generated by \texttt{Claude-3.7-Sonnet} and \texttt{GPT-4o-2024-11-20}.}
\resizebox{0.95\textwidth}{!}{
\label{tab:layout_comparison}
\begin{threeparttable}
\begin{tabular}{lcc|lcc}
\toprule
\textbf{Layout Statistic} & \textbf{Claude} & \textbf{GPT} & \textbf{Layout Statistic} & \textbf{Claude} & \textbf{GPT} \\
\midrule
\multicolumn{3}{l|}{\textbf{General}} & \multicolumn{3}{l}{\textbf{Tallest Column}} \\
Total columns & 376 & 293 & Height (px) & 7272.22 & 5794.42 \\
\multicolumn{3}{l|}{\textbf{Balance}} & Text length (char) & 2057.37 & 1554.44 \\
Relative position\tnote{1} & 0.55 & 0.51 & Number of images & 2.93 & 2.30 \\
Height coefficient of variation\tnote{2} & 0.21 & 0.18 & \multicolumn{3}{l}{\textbf{Shortest Column}} \\
Height ratio (max/min)\tnote{3} & 1.73 & 1.61 & Height (px) & 4379.82 & 3979.53 \\
Blank space proportion\tnote{4} & 19.16\% & 14.92\% & Text length (char) & 1392.26 & 1296.84 \\
 &  &  & Number of images & 1.74 & 1.59 \\
\bottomrule
\end{tabular}
\begin{tablenotes}\footnotesize
\item [1] Index of relative column positions within posters; values closer to 0.5 indicate more centered, balanced layouts.
\item [2] Measure of height consistency across columns; lower values indicate more uniform column heights.
\item [3] Ratio between tallest and shortest columns; values closer to 1 indicate more even column heights.
\item [4] Percentage of total poster area occupied by blank space.
\end{tablenotes}
\end{threeparttable}
}
\end{table}

\section{Related Work}

\paragraph{Poster Generation.} Academic poster generation involves creating a poster that summarizes the key information from an academic paper. 
Paramita et al. \cite{paramita2016tailored} develop a model that extracts essential sentences into templates to generate text-based posters.
Qiang et al. \cite{qiang2019learning} propose a more comprehensive method, decomposing poster generation into three subtasks: content extraction \cite{mihalcea2004textrank, xu2021neural, cheng2024sviptr,cheng2024xformparser}, panel attribute inference \cite{zhong2019publaynet,li2020docbank,huang2022layoutlmv3}, and panel layout generation \cite{lin2024layoutprompter,zhang2023layoutdiffusion}. 
Postdoc \cite{jaisankar2024postdoc} utilizes MLLMs to generate template-based posters but cannot produce flexible layouts with more dynamic integration of figures and text. Additionally, existing academic poster datasets \cite{yaoscipg,xu2021neural,qiang2016learning,wang2024scipostlayout,saxena2025postersum} lack fine-grained evaluation metrics necessary for comprehensive quality assessment.

\paragraph{HTML Code Generation and Multi-Agent.}
Recent research in automated front-end development focuses on generating HTML from diverse inputs such as screenshots, prototypes and natural language. This has spurred the creation of benchmarks like Design2Code~\citep{si2024design2code,xcoder}, Websight~\citep{laurenccon2024unlocking}, WebCode2M~\citep{gui2025webcode2m}, and Web2Code~\citep{yun2024web2code}. Code generation methodologies vary, including direct translation, structured approaches such as DCGen's~\citep{wan2024automatically} divide-and-conquer strategy and UICopilot's~\citep{gui2025uicopilot} hierarchical generation. 
Applications target mobile UIs~\citep{xiao2024prototype2code, zhou2024bridging}, multi-page websites~\citep{wan2024mrweb}, and web design~\citep{xiao2024interaction2code,li2024sketch2code,zhang2024nldesign}, with model fine-tuning~\citep{liang2024waffle} enhancing performance. Multi-agent systems are increasingly adopted to complex tasks\cite{han2024llm,liu2025roleagent}; for instance, agentic workflows can convert designs to code~\citep{ding2025frontend,islam2024mapcoder}, and some systems employ distinct agents for sub-tasks with iterative human feedback\citep{wang2024multimodal}.

\paragraph{LLM as a Judge.}
The use of LLMs as evaluators, termed ``LLM-as-a-Judge,'' is well-studied and has demonstrated high consistency with human judgment, with early work focusing on LLMs evaluating other LLMs, as seen in JudgeLM \cite{zhu2023judgelm,codearena}. Subsequent research introduced systems like AUTO-J \cite{li2023generative}, leveraging pairwise and single-response evaluations to achieve strong agreement with human assessments \cite{bai2023touchstone,li2023alpacaeval,execrepobench,li2023evaluating,sunrepofixeval,wmt2021,sun2025bitsai}. With the rise of MLLMs, their potential as evaluators in multimodal tasks is being explored, as traditional metrics often fail to capture the nuances of complex multimodal outputs \cite{antol2015vqa,liu2023hallusionbench,liu2023lost,liu2023visual,liu2023aligning}. To enhance LLM evaluation capabilities, techniques such as Chain-of-Thought \cite{wei2021finetuned,chusurvey,chai2025xcot,sun2024unicoder} and Training-free instruction following \cite{brown2020language,wei2021finetuned} have been proposed, addressing the need for more robust evaluators in both unimodal and multimodal contexts.

\section{Conclusion}
\label{sec:conclusion}

In conclusion, we present \method{}, a multi-agent framework for academic poster generation that effectively transforms research papers into visually coherent and informationally faithful posters. Our approach introduces three specialized agents with dedicated checker modules that enable iterative refinement through reflection mechanisms. To advance this field, we introduce \instruct{}, the first large-scale instruction dataset specifically designed for paper-to-poster transformation, comprising over 30,000 high-quality examples. Additionally, we establish \bench{}, a comprehensive benchmark with dual evaluation methodologies that systematically assesses poster quality across multiple dimensions. Experimental results demonstrate that \method{} produces posters that approach or sometimes exceed the quality of human-created examples, particularly when employing reasoning-enhanced LLMs. This work establishes a foundation for future research in automated academic communication tools, with promising implications for enhancing research accessibility and dissemination efficiency.

\bibliographystyle{plain}
\bibliography{main.bib}

\newpage
\appendix

\section{Examples of Poster Generation}

\begin{figure}[t]
    \centering
    \includegraphics[width=0.98\linewidth]{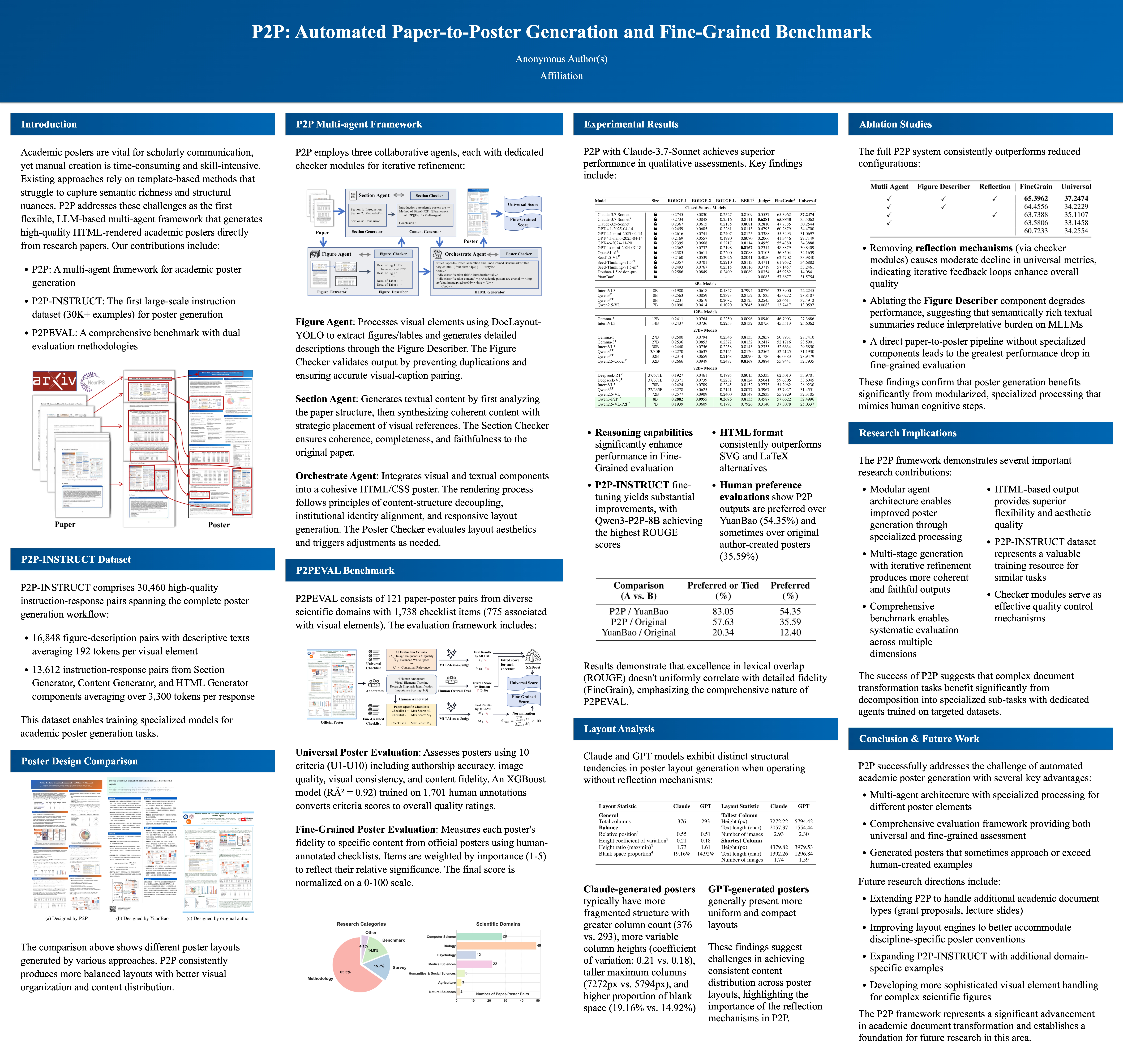}
    \caption{The poster for this paper, powered by \method{}.}    \label{fig:mine}
\end{figure}

\begin{figure}[h]
    \centering
    \begin{subfigure}[b]{0.32\textwidth}
        \centering
        \includegraphics[width=\textwidth]{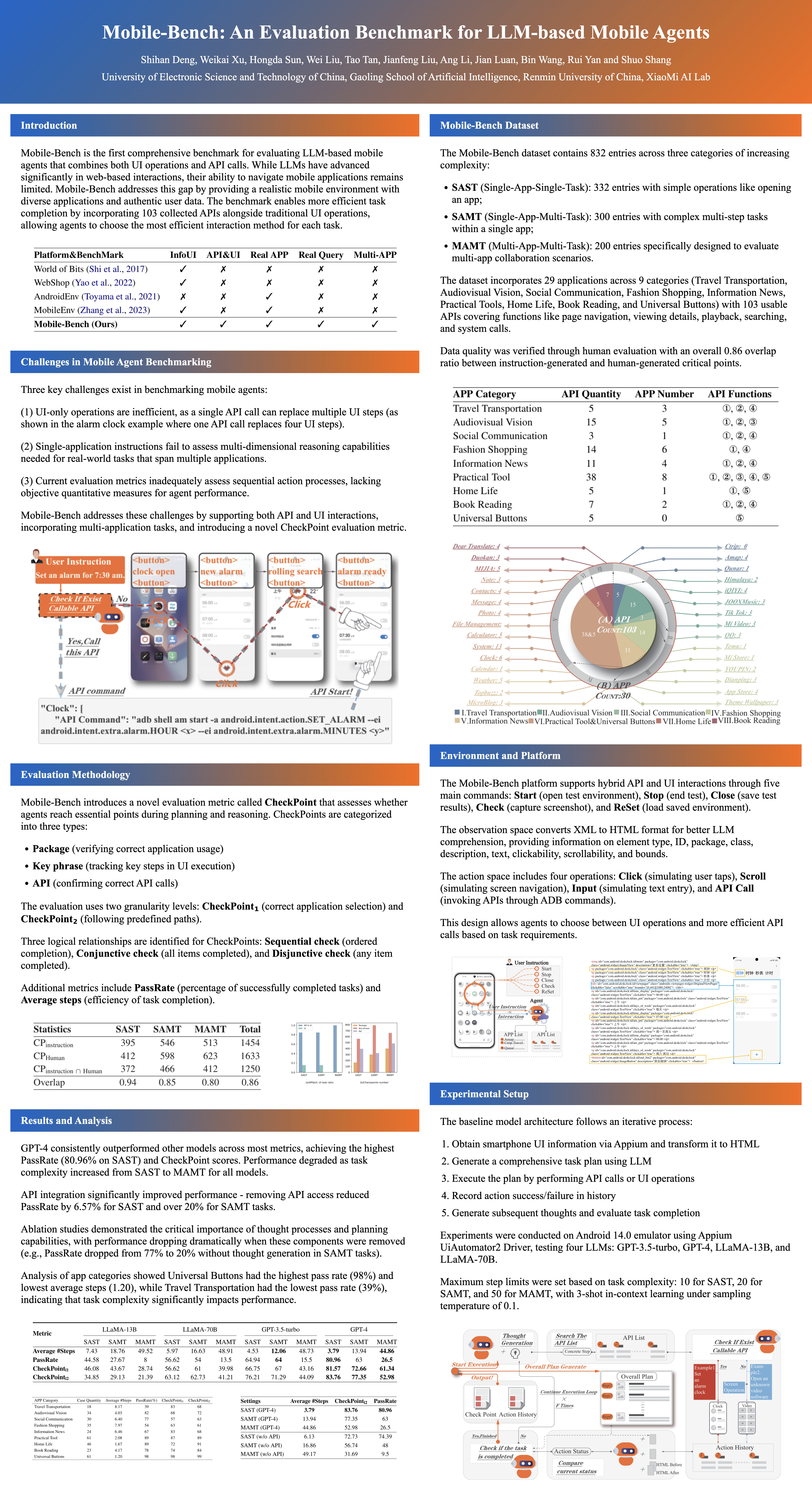}
        \caption{Designed by \model{}}
    \end{subfigure}
    \hfill
    \begin{subfigure}[b]{0.32\textwidth}
        \centering
        \includegraphics[width=\textwidth]{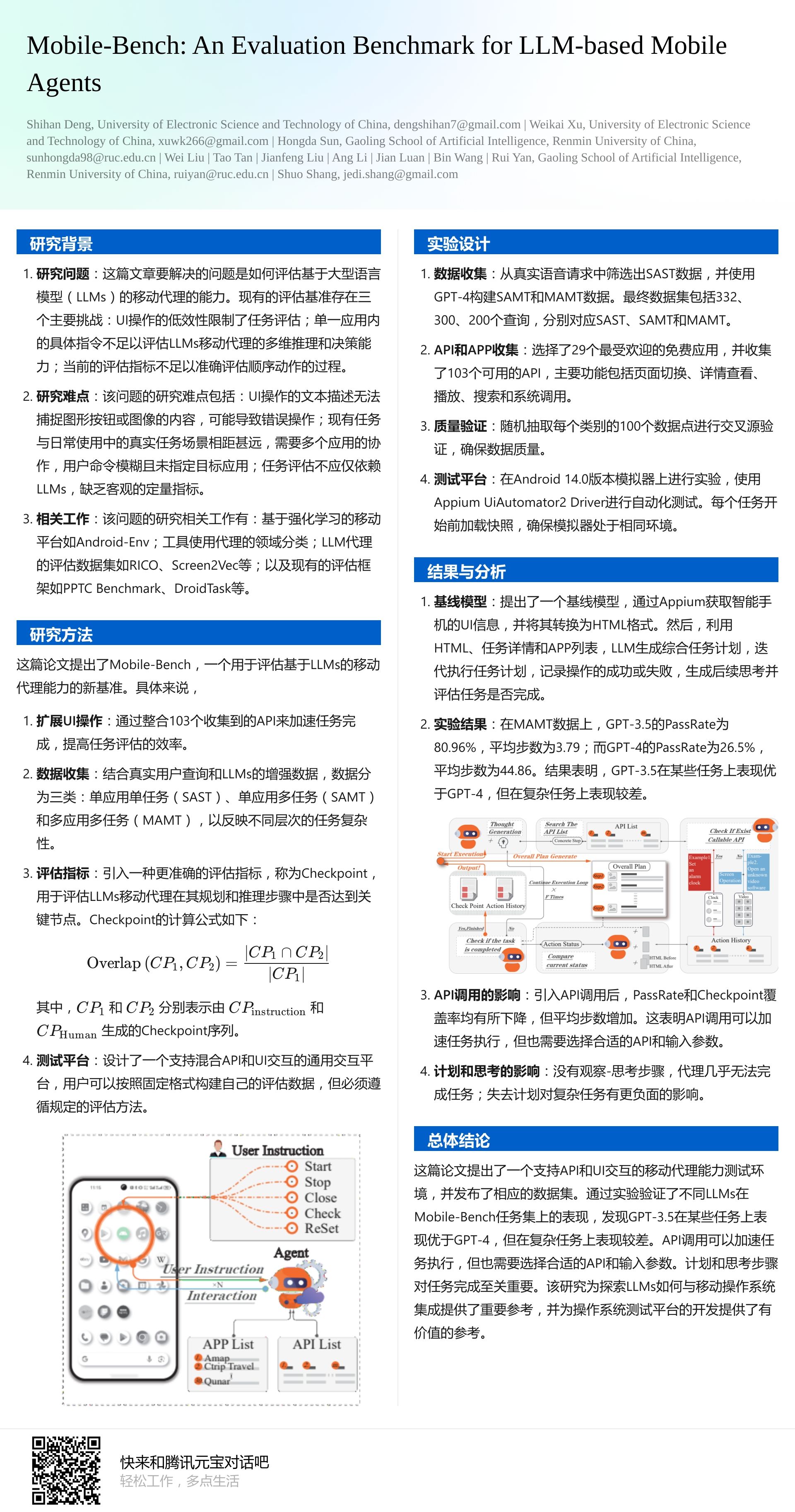}
        \caption{Designed by YuanBao}
    \end{subfigure}
    \hfill
    \begin{subfigure}[b]{0.32\textwidth}
        \centering
        \includegraphics[width=\textwidth]{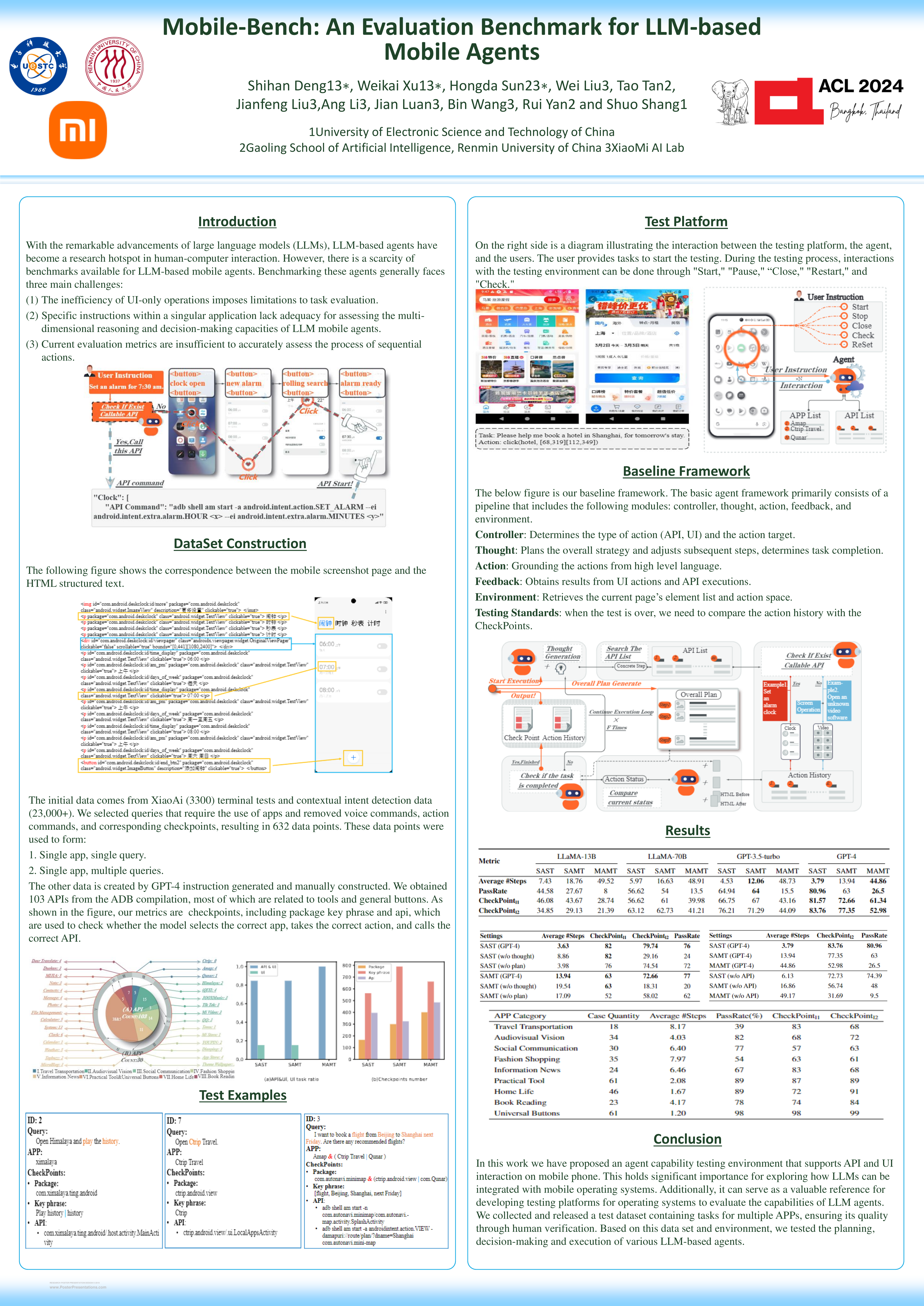}
        \caption{Designed by original author}
    \end{subfigure}
    \caption{Examples of academic poster design for \cite{deng2024mobile}.} 
    \label{fig:comparison}
\end{figure}

Examples of poster generation are shown in Fig~\ref{fig:mine} and Fig~\ref{fig:comparison}.

\section{The Features of Fine-Grained Poster Evaluation}
The Fine-Grained Poster Evaluation pipeline offers several distinct advantages: 
\begin{enumerate}
    \item \textbf{Ground-Truth Alignment}: Each checklist item references specific elements from the official academic posters and corresponding papers, ensuring accurate evaluation aligned with the original author's intent.
    \item \textbf{Domain-Specific Emphasis}: The pipeline captures domain-specific expectations and conventions, which universal criteria may overlook, reflecting discipline-specific priorities.
    \item \textbf{Essential Research Component Verification}: Critical content such as key figures, methodology details, and conclusions are explicitly accounted for using detailed scoring mechanisms, ensuring comprehensive evaluation.
    \item \textbf{Human Preference Integration}: Carefully calibrated by four human annotators, checklist item scores inherently encode domain expertise and human judgment regarding item significance and presentation quality.
\end{enumerate}

\section{The Features of HTML Format}
We compare the advantages of HTML for SVG and Latex:
\begin{itemize}
    \item Universal Accessibility and Portability: HTML posters can be easily viewed on any device with a web browser, requiring no specialized software (unlike LaTeX, which needs compilation, or potentially specific viewers for complex SVGs).
    \item  Rich Interactivity: HTML, often combined with CSS and JavaScript, allows for the seamless integration of interactive elements such as hyperlinks (to papers, datasets, author profiles), tooltips, expandable sections, or even embedded multimedia. This level of interactivity is more cumbersome to achieve and less natively supported in LaTeX or static SVG.
    \item  Flexible and Modern Styling: CSS provides powerful and flexible control over the visual presentation, enabling modern, responsive, and aesthetically engaging designs that can adapt to various screen sizes. This offers more design freedom than typical LaTeX layouts and better structural organization for complex content than a single SVG.
    \item  Ease of Web Integration: As the native language of the web, HTML posters can be effortlessly embedded into websites, shared via links, and are inherently well-suited for online conference platforms and digital dissemination.
\end{itemize}

\section{Prompt}

\definecolor{customblue}{HTML}{005bac}

\newtcolorbox[auto counter, number within=section]{bluebox}[2][]{%
  colback=white,
  colframe=customblue, 
  width=\textwidth, 
  arc=5mm, 
  boxrule=0.8mm, 
  title=\large #2, 
  breakable, 
  fonttitle=\small, 
  fontupper=\footnotesize, 
  #1 
}

\begin{bluebox}{Section Extraction}

\vspace{0.5em}

You are an expert in academic paper analysis.

\vspace{0.5em}

Please analyze the paper content and identify the sections that should be included in the poster. 

\vspace{0.5em}

For each section, provide a simple description of what should be included. First, determine the type of the paper. If it is a methodology research paper, focus on the method description, experimental results, and research methodology. If it is a benchmark paper, pay attention to task definitions, dataset construction, and evaluation outcomes. For survey/review papers, emphasize the significance of the field, key timelines or developmental milestones, critical theories and techniques, current challenges, and emerging trends. The above are just references; the specific section names should depend on the paper's content.

\vspace{0.5em}

Relevant sections for comparison can be combined. There should not be too many sections. The acknowledgement and references section should not appear.

\vspace{0.5em}

Return the result as a JSON object with section names as keys and descriptions as values. 

\vspace{0.5em}

Ensure the JSON is flat, without nested dictionaries or complex structures.

\vspace{0.5em}

\textbf{Example Format:}

\textit{(JSON Format Example.)}

    \vspace{0.5em}  

\textbf{Paper Content:}

\textit{(Content of Paper.)}

\end{bluebox}

\begin{bluebox}{Image Description}

\vspace{0.5em}

You are an academic image analysis expert. Your task is to provide detailed descriptions of academic figures, diagrams, charts, or images. Describe what the figure shows, its potential purpose in an academic paper, and any key data or trends visible. The description should be concise and to the point, and should not exceed 100 words.

\vspace{0.5em}

\textbf{Image Data:}

\textit{(Base64 PNG Image Data.)}

\end{bluebox}

\begin{bluebox}{Text-based Poster Generation}

\vspace{0.5em}

You are a helpful academic expert, who is specialized in generating a text-based paper poster, from given contents.

\vspace{0.5em}

\textbf{Figure Description:}

\textit{(Figures with Description.)}

\vspace{0.5em}  

\textbf{Paper Content:}

\textit{(Content of Paper.)}

\vspace{0.5em}

If the content of the poster can be described by figures, the relevant text-based content must be simplified to avoid redundancy. Important mathematical formulas can be appropriately placed to assist in understanding.

\vspace{0.5em}

All sections should be detailed in a markdown format. Do not use headings.

\end{bluebox}

\begin{bluebox}{Image-based Poster Generation}

\vspace{0.5em}

You are a helpful academic expert, who is specialized in generating a paper poster, from given contents and figures.

\vspace{0.5em}

\textbf{Figure Description:}

\textit{(Figures with Description.)}

\vspace{0.5em}  

\textbf{Text-based Poster:}

\textit{(Text-based Poster Content.)}

\vspace{0.5em}  

\textbf{Paper Content:}

\textit{(Content of Paper.)}

\vspace{0.5em}

Help me inside insert figures into my poster content using my figure index as `![figure\_description](figure\_index)`

\vspace{0.5em}

figure\_index starts from 0 and MUST be an integer, and don't use any other string in the figure\_index.

\vspace{0.5em}

Each figure can only be used once, and its placement should be precise and accurate.

\vspace{0.5em}

Use pictures and tables based on their importance.

\end{bluebox}

\begin{bluebox}{Poster Rendering}

\vspace{0.5em}

You are a professional academic poster web page creator and your task is to generate the HTML code for a nicely laid out academic poster web page based on the object provided.

\vspace{0.5em}

\textbf{Object Description:}

\begin{itemize}
    \item The object contains several fields. Each field represents a section, except for the title, author and affiliation fields. The field name is the title of the section and the field value is the Markdown content of the section.
    \item The image in Markdown is given in the format ![alt\_text, width = original\_width, height = original\_height, aspect ratio = aspect\_ratio](image\_index).
\end{itemize}

\vspace{0.5em}  

\textbf{HTML Structure:}

\begin{itemize}
    \item Only generate the HTML code inside <body>, without any other things.
    \item Do not use tags other than <div>, <p>, <ol>, <ul>, <li>, <img>, <strong>, <em>.
    \item Do not create sections that are not in the object.
    \item Place title, author and affiliation inside <div class="poster-header">. Place title inside <div class="poster-title">, author inside <div class="poster-author"> and affiliation inside <div class="poster-affiliation">.
    \item Place content inside <div class="poster-content">.
    \item Place each section inside <div class="section">. Place section title inside <div class="section-title"> and section content inside <div class="section-content">.
    \item Use <p> for paragraphs.
    \item Use <ol> and <li> for ordered lists, and <ul> and <li> for unordered lists.
    \item Use <img src="image\_index" alt="alt\_text"> for images.
\end{itemize}

\vspace{0.5em}  

\textbf{Color Specification:}

\begin{itemize}
    \item Do not add styles other than color, background, border, box-shadow.
    \item Do not add styles like width, height, padding, margin, font-size, font-weight, border-radius.
    \item Pick at least 2 colors from the visual identity of the affiliation. If there are multiple affiliations, consider the most well-known affiliation.
    \item For example, Tsinghua University uses \#660874 and \#d93379, Beihang University uses \#005bac and \#003da6, Zhejiang University uses \#003f88 and \#b01f24. These are just examples, you must pick colors from the visual identity of the affiliation.
    \item Add text and background color to poster header and section title using inline style. Use gradient to make the poster more beautiful.
    \item The text and background color of each section title should be the same.
\end{itemize}

\vspace{0.5em}

\textbf{Layout Specification:}

\begin{itemize}
    \item Optionally, inside <div class="poster-content">, group sections into columns using <div style="display: flex; gap: 1rem"> and <div class="poster-column" style="flex: 1">.
    \item You must determine the number and flex grow of columns to make the poster more balanced. If the height of one column is too large, move some sections into other columns.
    \item Optionally, inside <div class="section-content">, group texts and images into columns using <div style="display: flex; gap: 0.5rem"> and <div class="section-column" style="flex: 1">.
    \item For example, if there are two images in two columns whose aspect ratios are 1.2 and 2 respectively, the flex grow of two columns should be 1.2 and 2 respectively, to make the columns have the same height.
    \item Calculate the size of each image based on columns and aspect ratios. Add comment <!-- width = display\_width, height = display\_height --> before each image.
    \item Rearrange the structure and order of sections, texts and images to make the height of each column in the same group approximately the same.
    \item For example, if there are too many images in one section that make the height of the column too large, group the images into columns.
    \item DO NOT LEAVE MORE THAN 5\% BLANK SPACE IN THE POSTER.
\end{itemize}

\vspace{0.5em}

\textbf{Existing Style:}

\textit{(Existing CSS Style.)}

\vspace{0.5em}

\textbf{Object:}

\textit{(Poster Object.)}

\end{bluebox}

\end{CJK*}
\end{document}